\documentclass{ws-procs11x85}
\usepackage{ws-procs-thm}           

\usepackage{multirow}
\usepackage{makecell}
\usepackage{booktabs}
\usepackage{algorithm}
\usepackage{algpseudocode}
\DeclareMathOperator*{\argmax}{arg\,max}

\usepackage{float}
\usepackage{colortbl}
\usepackage{color}
\usepackage{soul}
\definecolor{lightred}{rgb}{1, 0.8, 0.8}
\definecolor{lightblue}{rgb}{0.8, 0.9, 1}

\newcommand*\colourcheck[1]{%
  \expandafter\newcommand\csname #1check\endcsname{\textcolor{#1}{\ding{52}}}%
}
\colourcheck{blue}
\colourcheck{green}
\colourcheck{red}
\usepackage{pifont}
\newcommand{\xmark}{\ding{55}}%

\usepackage{xcolor}

\begin{document}

\title{Improving Retrieval-Augmented Generation in Medicine with Iterative Follow-up Questions}

\author{Guangzhi Xiong$^{1,*}$, Qiao Jin$^{2,*}$, Xiao Wang$^3$, Minjia Zhang$^3$, Zhiyong Lu$^{2,\dag}$, Aidong Zhang$^{1,\dag}$}

\address{$^1$Department of Computer Science, University of Virginia,
VA 22904, USA\\
$^2$National Library of Medicine, National Institutes of Health, MD 20892, USA\\
$^3$Department of Computer Science, University of Illinois Urbana-Champaign, IL 61801, USA\\
$^*$Equal contribution. $^\dag$Co-correspondence.\\
E-mail: hhu4zu@virginia.edu, qiao.jin@nih.gov, xiaow4@illinois.edu, minjiaz@illinois.edu, zhiyong.lu@nih.gov, aidong@virginia.edu
}

\begin{abstract}
The emergent abilities of large language models (LLMs) have demonstrated great potential in solving medical questions. They can possess considerable medical knowledge, but may still hallucinate and are inflexible in the knowledge updates. While Retrieval-Augmented Generation (RAG) has been proposed to enhance the medical question-answering capabilities of LLMs with external knowledge bases, it may still fail in complex cases where multiple rounds of information-seeking are required. To address such an issue, we propose iterative RAG for medicine (\textit{i}-MedRAG), where LLMs can iteratively ask follow-up queries based on previous information-seeking attempts. In each iteration of \textit{i}-MedRAG, the follow-up queries will be answered by a conventional RAG system and they will be further used to guide the query generation in the next iteration. Our experiments show the improved performance of various LLMs brought by \textit{i}-MedRAG compared with conventional RAG on complex questions from clinical vignettes in the United States Medical Licensing Examination (USMLE), as well as various knowledge tests in the Massive Multitask Language Understanding (MMLU) dataset. Notably, our zero-shot \textit{i}-MedRAG outperforms all existing prompt engineering and fine-tuning methods on GPT-3.5, achieving an accuracy of 69.68\% on the MedQA dataset. In addition, we characterize the scaling properties of \textit{i}-MedRAG with different iterations of follow-up queries and different numbers of queries per iteration. Our case studies show that \textit{i}-MedRAG can flexibly ask follow-up queries to form reasoning chains, providing an in-depth analysis of medical questions. To the best of our knowledge, this is the first-of-its-kind study on incorporating follow-up queries into medical RAG. The implementation of \textit{i}-MedRAG is available at \url{https://github.com/Teddy-XiongGZ/MedRAG}.

\end{abstract}


\copyrightinfo{\copyright\ 2024 The Authors. Open Access chapter published by World Scientific Publishing Company and distributed under the terms of the Creative Commons Attribution Non-Commercial (CC BY-NC) 4.0 License.}


\newpage
\section{Introduction}

Generative artificial intelligence (AI) technologies such as large language models (LLMs) have brought a wide variety of opportunities for biomedical applications \cite{thirunavukarasu2023large, zhou2023survey, omiye2024large, tian2024opportunities}. 
For example, they have shown great potential for answering biomedical questions \cite{luo2022biogpt, lievin2024can, nori2023can, singhal2023large, bolton2024biomedlm}, summarizing medical documents \cite{shaib2023summarizing, tang2023evaluating, van2024adapted}, and matching patients to clinical trials \cite{jin2023matching, wong2023scaling, wornow2024zero, zhuang2024team}.
However, LLMs often generate plausible-sounding but inaccurate content, an issue commonly known as ``hallucination'' in the literature\cite{ji2023survey}.
They also possess outdated knowledge obtained from a fixed set of training data \cite{wu2024continual}.
Retrieval-augmented generation (RAG) provides a lightweight post-training solution to these issues by providing LLMs with relevant documents retrieved from up-to-date and trustworthy sources \cite{lewis2020retrieval, gao2023retrieval}.

While there have been several medical applications of RAG, such as Almanac \cite{zakka2024almanac}, Clinfo.ai \cite{lozano2023clinfo}, and MedRAG \cite{xiong2024benchmarking}, their RAG component is mainly beneficial to questions that have direct answers in a single document, such as those in the PubMedQA \cite{jin2019pubmedqa} and BioASQ \cite{tsatsaronis2015overview} datasets. 
However, only marginal improvements are seen with RAG for questions that require multiple rounds of clinical reasoning like MedQA \cite{jin2021disease}, a dataset curated from medical license examinations.
For example, to recommend a treatment for a patient with certain symptoms, a system needs to first infer the potential diagnosis from the symptoms and then find a suitable treatment for the diagnosis.
Nevertheless, only one round of retrieval is conducted in the conventional RAG architecture, prohibiting multiple rounds of information seeking that are required in complex clinical reasoning.

In this work, we propose \textit{i}-MedRAG, a simple and effective framework for incorporating follow-up queries into RAG.
Specifically, we prompt LLMs to iteratively generate follow-up queries to search for additional information from external medical corpora. The queries and the corresponding answers generated with RAG will be used to augment the answer generation of the original question. 
Empirical results demonstrate the effectiveness of \textit{i}-MedRAG on both open- and close-source LLMs, which show improved performance on the United States Medical Licensing Examination (USMLE) subset of MedQA and medical questions from the Massive Multitask Language Understanding (MMLU) dataset.
Our further analysis of the number of iterations and number of queries per iteration used in \textit{i}-MedRAG reflects how its performance scales with different settings.
Additionally, we present several case studies of \textit{i}-MedRAG, showing how it overcomes the limitations in conventional RAG to find the correct answers.

In summary, our contributions are three-fold:
\begin{itemize}
    \item We introduce \textit{i}-MedRAG, a novel RAG architecture that incorporates follow-up queries to solve complex reasoning tasks.
    \item We have conducted comprehensive experiments on medical question answering, and the results demonstrate that \textit{i}-MedRAG not only outperforms conventional RAG approaches but also surpasses all other prompt engineering approaches on MedQA with GPT-3.5, setting a new state-of-the-art performance of 69.68\%.
    \item We also provide analyses to further characterize \textit{i}-MedRAG, showing how its performance varies with the scaling of follow-up queries.
\end{itemize}

\section{Related Work}

\subsection{Retrieval-Augmented Generation for Medicine}
Retrieval-augmented generation (RAG) has been widely adopted in medicine.
Here, we discuss several representative approaches.
Almanac \cite{zakka2024almanac} is a system that augments LLMs with curated resources for medical guidelines and treatment recommendations, which shows improvements over the standard LLMs in six manually assessed metrics.
Similarly, Low \textit{et al.} \cite{low2024answering} demonstrate the improvements of RAG-based systems for real-world clinical queries with manual evaluation.
Clinfo.ai \cite{lozano2023clinfo} is an open-source web application that answers clinical questions based on retrieved scientific literature from PubMed articles.
Xiong \textit{et al.} \cite{xiong2024benchmarking} conduct a benchmarking study with the MedRAG toolkit, and show the benefits of RAG in several medical multi-choice question answering datasets.
There are also various biomedical literature search products \cite{jin2024pubmed} that use RAG to summarize the retrieved articles \cite{jin2023retrieve}, such as OpenEvidence\footnote{\url{https://www.openevidence.com/}} and ChatRWD\footnote{\url{https://www.atroposhealth.com/chatrwd}}.
However, most of the RAG studies in medicine use the conventional architecture with only one round of retrieval.
There have been several attempts to use iterative data refinement for LLM training\cite{li2024self} or RAG\cite{shao2023enhancing,trivedi2023interleaving,jiang2024retrieve} in the general domain. Nevertheless, similar ideas have not yet been explored in the medical domain.
To the best of our knowledge, our study presents the first approach and evaluations on incorporating follow-up queries in RAG for medicine.

\begin{figure}
    \centering
    \includegraphics[width=0.618\linewidth]{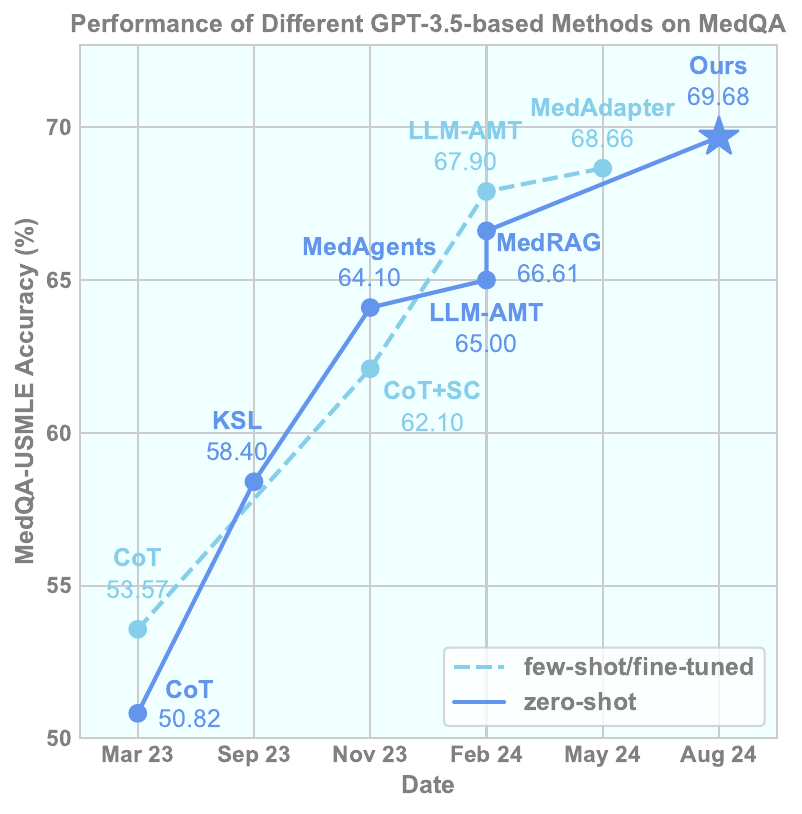}
    \caption{Comparison of various methods proposed to improve GPT-3.5 performance on MedQA. Our zero-shot \textit{i}-MedRAG outperforms all previous prompt engineering and fine-tuning methods.}
    \label{fig:medqa_performance}
\end{figure}

\subsection{Medical Question Answering}
Question answering tasks such as MedQA \cite{jin2021disease}, PubMedQA \cite{jin2019pubmedqa}, MedMCQA \cite{pal2022medmcqa}, BioASQ \cite{tsatsaronis2015overview}, and Massive Multitask Language Understanding (MMLU) \cite{hendrycks2020measuring} are commonly used to benchmark the medical knowledge and reasoning capabilities of LLMs \cite{jin2022biomedical}.
Most of these datasets focus on single-hop questions such as ``what is the most common symptom of hypertension?'', while only MedQA questions are longer patient vignettes where both medical knowledge and multi-step reasoning are required.
As such, there have been many studies working on improving the GPT-3.5 performance on MedQA with prompt engineering. 
Figure~\ref{fig:medqa_performance} shows the comparison among different representative prompt engineering approaches on MedQA, including chain-of-thought (CoT) prompting \cite{wei2022chain}, self-consistency (SC) prompting \cite{wang2022self}, multi-agent communication with MedAgents \cite{tang2023medagents}, and RAG-based approaches such as Knowledge Solver (KSL) \cite{feng2023knowledge},  LLMs Augmented with Medical Textbooks (LLM-AMT) \cite{wang2023augmenting}, and MedRAG \cite{xiong2024benchmarking}.
Much fewer studies focus on prompt engineering with GPT-4 on MedQA \cite{nori2023can, savage2024diagnostic}, probably because the raw GPT-4 error rate \cite{nori2023capabilities} is close to the noise rate in MedQA annotations \cite{saab2024capabilities}.
In this study, we focus on the zero-shot setting as it reflects realistic clinical scenarios.
While not requiring any instances for training or few-shot learning, our approach surpasses all previous methods with GPT-3.5 on the MedQA dataset.

\section{Methods}

\begin{figure}
    \centering
    \includegraphics[width=1\linewidth]{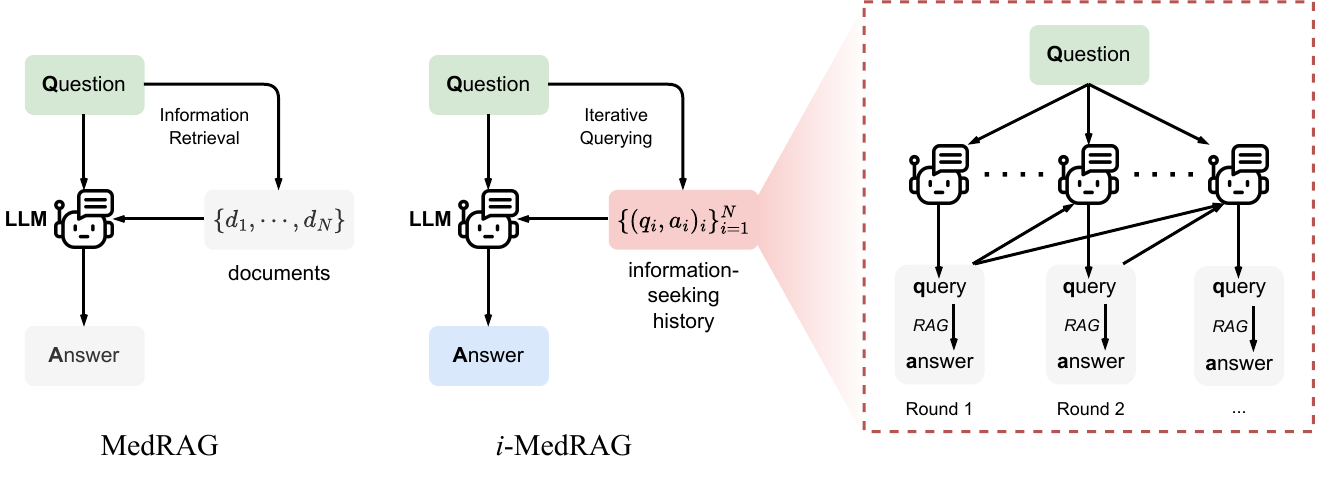}
    \caption{Overview of \textit{i}-MedRAG and its comparison to RAG (MedRAG). 
    Left: the pipeline of Retrieval-Augmented Generation (RAG).
    Middle: the pipeline of our proposed \textit{i}-MedRAG.
    Right: the iterative generation of question-specific medical query-answer (QA) pairs by asking follow-up queries.}
    \label{fig:overview}
\end{figure}

Figure \ref{fig:overview} shows the overview of our \textit{i}-MedRAG and its comparison to the conventional Retrieval-Augmented Generation (RAG).
Different from RAG, our \textit{i}-MedRAG modifies its pipeline by replacing the information retrieval step (Figure 2 left) with our proposed iterative question-answering step (Figure 2 middle and right). The settings of RAG are described in Section \ref{sec:rag} and the pipeline of our new \textit{i}-MedRAG is discussed in Section \ref{sec:iter-RAG}. The details of the iterative question answering are described in Section \ref{sec:questioning}.

\subsection{Retrieval-Augmented Generation} \label{sec:rag}

In the zero-shot setting of medical question answering, the task of LLM \(\mathcal{M}\) is trying to find the correct answer \(\mathcal{A}\) given the question \(\mathcal{Q}\) only. The ideal answer prediction \(\Tilde{\mathcal{A}}\) can be provided by
\begin{equation}
    \Tilde{\mathcal{A}} = \argmax_{\mathcal{A}} \mathbb{P}_{\mathcal{M}}(\mathcal{A}\mid \mathcal{Q}, \text{inst.}),
\end{equation}
where the ``inst.'' is the task instruction the user provides that instructs the model to perform the task. As medical questions are knowledge-intensive \cite{jin2022biomedical}, it benefits from accessing large-scale external corpora to search for useful information \cite{zakka2024almanac,lozano2023clinfo,xiong2024benchmarking}. A typical method to combine LLM reasoning with external corpora is RAG, which first retrieves relevant documents from the corpus for the given medical question and enters the retrieved documents along with the question into LLM to augment its answer generation. Formally, the RAG pipeline can be described as
\begin{equation}\label{eq:rag}
    \Tilde{\mathcal{A}} = \textit{RAG}(\mathcal{Q}; \mathcal{M}, \mathcal{R}, \mathcal{D}) = \argmax_{\mathcal{A}} \mathbb{P}_{\mathcal{M}}(\mathcal{A}\mid \mathcal{Q}, \text{inst.}, \{d_i\}_{i=1}^N),
\end{equation}
where \(\{d_i\}_{i=1}^N\) are the question-specific retrieved documents given by
\begin{equation}\label{eq:retrieval}
    \{d_i\}_{i=1}^N = \mathcal{R}(\mathcal{Q}; \mathcal{D}).
\end{equation}
Here \(R\) is the text retriever and \(\mathcal{D}\) is the corpus with a collection of documents. 

\subsection{Iterative Retrieval-Augmented Generation} \label{sec:iter-RAG}

While RAG exhibits promising performance in medical question answering \cite{xiong2024benchmarking}, it may be unable to handle certain complex medical questions in real-world cases. 
As text retrievers are typically trained to find relevant documents based on text similarity or lexicon overlap, they cannot break down a complex question and search for relevant information in a step-by-step manner.
Thus, the inflexible retrieval step (Formula \ref{eq:retrieval}) in RAG may fail to analyze medical questions and find useful information to augment the answer generation, especially in complex clinical cases, where multiple rounds of information-seeking are required. 

To address the issues mentioned, we propose to incorporate flexible information retrieval by 
prompting LLMs to iteratively generate follow-up queries based on the given medical question and previous information-seeking history.
Moreover, as the context lengths of LLMs are limited, it can be impractical and infeasible to include all retrieved documents in the LLM context. Therefore, we prompt LLMs to directly answer the raised queries with relevant information and use such query-answer pairs as the information-seeking history.
The pipeline of our proposed system can be formulated as
\begin{equation}
    \Tilde{\mathcal{A}} = \textit{\textit{i}-MedRAG}(\mathcal{Q}; \mathcal{M}, \mathcal{R}, \mathcal{D})  = \argmax_{\mathcal{A}} \mathbb{P}_{\mathcal{M}}(\mathcal{A}\mid \mathcal{Q}, \text{inst.}, \{(q_i, a_i)\}_{i=1}^N),
\end{equation}
where \(\{(q_i, a_i)\}_{i=1}^N\) are the queries and the corresponding answers generated by LLMs with the help of RAG. The iterative process of query and answer generation will be detailed in Section \ref{sec:questioning}.

\subsection{Iterative Generation of Follow-up Questions} \label{sec:questioning}

While the retrieved documents in RAG are determined by the question and the retrieval system, we propose to incorporate the reasoning capabilities of LLMs in \textit{i}-MedRAG by prompting them to dynamically generate helpful queries in a step-by-step manner. Specifically, the LLM will be encouraged to generate \(n\) different queries to help find useful additional information for \(m\) iterations. In all iterations except for the first one, the model will be given the information-seeking history to generate context-specific follow-up queries. The queries \(q_{i1}, \cdots, q_{in}\) generated in the \(i\)-th iteration can be formulated as
\begin{equation}
    q_{i1}, \cdots, q_{in} = \left\{
    \begin{aligned}
        &\argmax_{q_{i1}, \cdots, q_{in}} \mathbb{P}_{\mathcal{M}}(q_{i1}, \cdots, q_{in} \mid \mathcal{Q}, \text{inst.}'), &\text{if }i=1, \\
        &\argmax_{q_{ic1}, \cdots, q_{in}} \mathbb{P}_{\mathcal{M}}(q_{i1}, \cdots, q_{in} \mid \mathcal{Q}, \text{inst.}', \{(q_{jk}, a_{jk})\}_{j=1, \cdots, i-1}^{k=1, \cdots, n}), &\text{if }i>1.
    \end{aligned}
    \right.
\end{equation}

Different from the ``inst.'' in Formula \ref{eq:rag}, the ``inst.$'$'' here is a modified instruction which focuses on generating follow-up queries instead of answering the medical question. For each query generation step, we prompt the LLM to analyze the existing information first and then generate new queries for additional knowledge. The step-by-step ``reason-then-query'' pipeline helps LLMs break down complex medical questions and find useful information from the external corpus.
The answer to each generated query is given by a RAG system mentioned in Formula \ref{eq:rag}. This enables the system to leverage existing literature to provide grounded answers for generated queries.

The overall algorithm of \textit{i}-MedRAG is presented in Algorithm \ref{alg:algorithm}.

\begin{algorithm*}
\caption{The algorithm of \textit{i}-MedRAG for medical question answering}
\textbf{Input} medical question $\mathcal{Q}$, large language model $\mathcal{M}$, text retriever $\mathcal{R}$, medical corpus $\mathcal{D}$, query instruction ``inst.$'$'', answer instruction ``inst.'', hyperparameters \(m, n, N\) \\
\textbf{Output} answer prediction $\Tilde{\mathcal{A}}$
\begin{algorithmic}[1]
\State Initialize the information-seeking history $\mathcal{H} = $ emptylist()
\For{$i$ in $1,2,\cdots,m$}
    \If{$i = 1$}
    \State generate \(n\) new queries \(q_{i1}, \cdots, q_{in}\) using \(\mathcal{M}\) given \(\mathcal{Q}\)
    \ElsIf{$i > 1$}
    \State generate \(n\) new queries \(q_{i1}, \cdots, q_{in}\) using \(\mathcal{M}\) given \(\mathcal{Q}\) and \(\mathcal{H}\)
    \EndIf
    \For{$j$ in $1,2,\cdots,n$}
        \State retrieve \(N\) relevant documents \(d_{ij}^1, \cdots, d_{ij}^N\) using \(\mathcal{R}\) and \(\mathcal{D}\) given \(q_{ij}\)
        \State
        generate the answer \(a_{ij}\) using \(\mathcal{M}\) given \(q_{ij}\) and \(d_{ij}^1, \cdots, d_{ij}^N\)
        \State add the query-answer pair \((q_{ij}, a_{ij})\) to the list $\mathcal{H}$
    \EndFor
\EndFor
\State generate the predicted answer $\Tilde{\mathcal{A}}$ using \(\mathcal{M}\) given \(\mathcal{Q}\) and \(\mathcal{H}\)
\State \textbf{return} $\Tilde{\mathcal{A}}$
\end{algorithmic}
\label{alg:algorithm}
\end{algorithm*}
\section{Experiments}

\subsection{Evaluation settings}

To evaluate the performance of our proposed \textit{i}-MedRAG on knowledge-intensive medical question-answering tasks and compare it with other approaches, we select MedQA \cite{jin2021disease} as the testbed, which contains medical questions collected from United States Medical Licensing Examination (USMLE). With complex clinical cases in the dataset, MedQA reflects the difficulty of decision-making in real-world clinical medicine. The approaches for comparison are prompt engineering or fine-tuning methods that try to improve the performance of GPT-3.5 on MedQA, including chain-of-thought (CoT) prompting\cite{nori2023capabilities}, self consistency (SC), knowledge solver (KSL)\cite{feng2023knowledge}, medical agents (MedAgents)\cite{tang2023medagents}, LLMs augmented with medical textbooks (LLM-AMT)\cite{wang2023augmenting}, medical retrieval-augmented generation (MedRAG)\cite{xiong2024benchmarking}, and LLMs with test-time adaptations (MedAdapter)\cite{shi2024medadapter}. 

Additionally, we evaluate the generalizability of our \textit{i}-MedRAG with more LLMs and medical datasets. Llama-3.1-8B is selected as the representative of open-source models, which has a context window of 128k tokens. We also include MMLU-Med, a set of six medical tasks (anatomy, clinical knowledge, professional medicine, human genetics, college medicine, college biology) from Massive Multitask Language Understanding (MMLU), following previous studies \cite{singhal2023large,xiong2024benchmarking}. 
MMLU-Med serves as a testbed to show the performance of \textit{i}-MedRAG on a variety of different medical tasks.

Both MedQA and MMLU-Med are composed of multi-choice questions, whose evaluation metric is the accuracy of predicted answers chosen from given options.
For the retrieval part in \textit{i}-MedRAG, we select the Textbooks\cite{jin2021disease} and Statpearls\footnote{\url{https://www.statpearls.com/}} corpora introduced in MedRAG\cite{xiong2024benchmarking}, which are shown effective on medical examination questions. MedCPT\cite{jin2023medcpt} is chosen as the text retriever, which has been trained on domain-specific literature. For other baselines compared, the official settings described in their papers are used.

\begin{table}[h!]
    \tbl{Performance of GPT-3.5 with different prompt engineering / fine-tuning methods on MedQA. The ``External Knowledge'' column denotes if the method augments LLM generation with information retrieval of external knowledge.}{
    \centering
    \begin{tabular}{lcccccccccccc}
        \toprule
        Method & External Knowledge & Setting & Accuracy (\%) \\
        \midrule
        Chain of Thought \cite{nori2023capabilities} & No & zero-shot & 50.82 \\
        Knowledge Solver \cite{feng2023knowledge} & Yes & zero-shot & 58.40 \\
        Chain of Thought + Self Consistency \cite{tang2023medagents} & No & zero-shot & 61.30 \\
        MedAgents\cite{tang2023medagents} & No & zero-shot & 64.10 \\
        LLMs Augmented with Medical Textbook\cite{wang2023augmenting} & Yes & zero-shot & 65.00 \\
        MedRAG\cite{xiong2024benchmarking} & Yes & zero-shot & 66.61 \\
        \midrule
        Chain of Thought \cite{nori2023capabilities} & No & five-shot & 53.57 \\
        Chain of Thought + Self Consistency \cite{tang2023medagents} & No & five-shot & 62.10 \\
        LLMs Augmented with Medical Textbook \cite{wang2023augmenting} & Yes & fine-tuned & 67.90 \\
        MedAdapter\cite{shi2024medadapter} & No & fine-tuned & 68.66 \\
        \midrule
        \bf \textit{i}-MedRAG (ours) & \bf Yes & \bf zero-shot & \bf 69.68 \\
        \bottomrule
    \end{tabular}}
    \label{tab:performance_gpt}
\end{table}

\subsection{Main results}

Table \ref{tab:performance_gpt} shows the comparison results of \textit{i}-MedRAG and other baseline approaches on MedQA using GPT-3.5. 
Official scores reported in previous research are used for a fair comparison.
While methods with few-shot learning or model fine-tuning tend to perform better than the ones in a zero-shot setting, our \textit{i}-MedRAG set a state-of-the-art performance of GPT-3.5 on MedQA without any training samples or parameter tuning. Among zero-shot approaches, \textit{i}-MedRAG (69.68\%) has a significant performance improvement (\(p<0.05\)) compared to the previous best record achieved by MedRAG (66.61\%).

The results of generalizing \textit{i}-MedRAG to more LLMs and data are presented in Table \ref{tab:performance_generalizability}. We compare \textit{i}-MedRAG with our implemented CoT and MedRAG to see if \textit{i}-MedRAG can bring a consistent improvement of LLM performance in medical question answering.
For all experiments with \textit{i}-MedRAG, we tune the hyperparameters on a validation set of 100 samples and then report its scores on the test set.
Similar to the results on GPT-3.5, the open-source Llama-3.1-8B also shows improved performance on MedQA with the help of \textit{i}-MedRAG. While Llama-3.1-8B shows a close performance to GPT-3.5 in CoT and MedRAG settings, its performance is significantly improved with \textit{i}-MedRAG, achieving an accuracy of 75.02\%. The improved performance of GPT-3.5 and Llama-3.1-8B on MMLU-Med also demonstrates the generalizability of \textit{i}-MedRAG to more medical data. As medical questions in MMLU-Med are less complex than the USMLE questions in MedQA, follow-up queries may not be necessary to find relevant information for the given question. Thus, it can be observed that the improvement by \textit{i}-MedRAG compared to MedRAG is less significant in MMLU-Med than in MedQA.

\begin{table}\small
    \tbl{Performance of \textit{i}-MedRAG on different LLMs and datasets. ``Acc.'' denotes the accuracy. ``\(\Delta\)'' shows the relative performance improvement compared with CoT.}{
    \centering
    \begin{tabular}{lccccccccccc}
        \toprule
        \bf \multirow{3}{*}{\bf Model} & \bf \multirow{3}{*}{\bf Method} & \multicolumn{2}{c}{\bf MedQA-USMLE} & \multicolumn{2}{c}{\bf MMLU-Med} & \multicolumn{2}{c}{\bf Average} \\
        \cmidrule(lr){3-4} \cmidrule(lr){5-6} \cmidrule(lr){7-8}
        & & Acc. & \(\Delta\) & Acc. & \(\Delta\) & Acc. & \(\Delta\) \\
        \midrule
        GPT-3.5-Turbo & CoT & 65.04 & +0.00\% & 72.91 & +0.00\% & 68.98 & +0.00\% \\
        GPT-3.5-Turbo & MedRAG & 66.61 & +2.41\% & 75.48 & +3.52\% & 71.05 & +3.00\% \\
        GPT-3.5-Turbo & \textit{i}-MedRAG & \textbf{69.68} & \textbf{+7.13\%} & \textbf{75.85} & \textbf{+4.03\%} & \textbf{72.77} & \textbf{+5.49\%} \\
        \midrule
        Llama-3.1-8B & CoT & 64.73 & +0.00\% & 77.23 & +0.00\% & 70.98 & +0.00\% \\
        Llama-3.1-8B & MedRAG & 66.54 & +2.80\% & 78.05 & +1.06\% & 72.30 & +1.86\% \\
        Llama-3.1-8B & \textit{i}-MedRAG & \textbf{73.61} & \textbf{+13.72\%} & \textbf{78.42} & \textbf{+1.54\%} & \textbf{76.02} & \textbf{+7.10\%} \\
        \bottomrule
    \end{tabular}}
    \label{tab:performance_generalizability}
\end{table}

\subsection{Scaling with iterations and queries}

As we described in Section \ref{sec:questioning}, the number of iterations to ask follow-up queries and the number of queries generated in each iteration are the two critical hyperparameters in our proposed iterative generation of follow-up queries. To explore how different selections of the hyperparameter values affect the model performance, we run \textit{i}-MedRAG with different settings and compare their results. We test both GPT-3.5 and Llama-3.1-8B on MedQA and MMLU-Med to examine if there are model-specific or task-specific patterns. 

\begin{figure}
    \centering    \includegraphics[width=\linewidth]{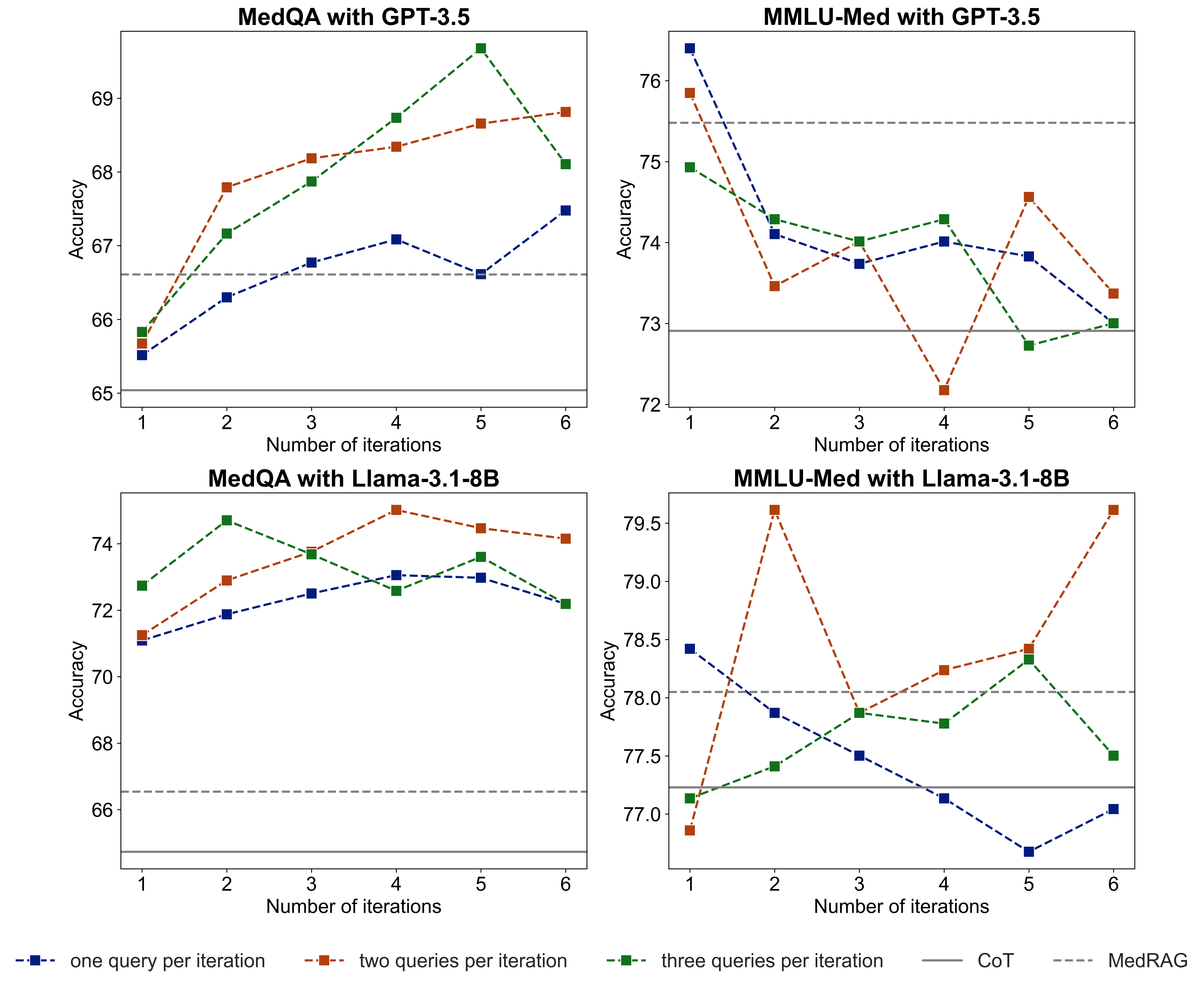}
    \caption{\textit{i}-MedRAG performance on MedQA and MMLU-Med with different hyperparameters.}
    \label{fig:scaling}
\end{figure}

Figure \ref{fig:scaling} shows the model performance with different hyperparameter settings. Generally, MedQA and MMLU-Med show distinct patterns in performance change with the increasing number of iterations. While the performance of both GPT-3.5 and Llama-3.1-8B on MedQA tends to improve with more iterations of follow-up queries, their performance on MMLU-Med converges or starts to drop with just one or two iterations, corresponding to the different complexities of these two tasks.

From the results on MedQA, it is also empirically shown that the number of generated queries per iteration determines the rate of performance improvement and convergence over multiple iterations. LLMs with more queries generated per iteration tend to have a larger improvement in accuracy but also converge more quickly. Such a result is intuitively reasonable as more information can be collected each iteration with more generated queries.

\subsection{Case studies}

\begin{table}[h!] \small
    \tbl{An example of how \textit{i}-MedRAG helps GPT-3.5 to find the correct answer. The \sethlcolor{lightred}\hl{incorrect rationale} and \sethlcolor{yellow}\hl{useful evidence} are highlighted.}{
    \centering
    \begin{tabular}{p{16.5cm}}
        \toprule
        \bf \makecell[c]{MedQA Question} \\
        \midrule
\textbf{0001}: A 67-year-old man with transitional cell carcinoma of the bladder comes to the physician because of a 2-day history of ringing sensation in his ear. He received this first course of neoadjuvant chemotherapy 1 week ago. Pure tone audiometry shows a sensorineural hearing loss of 45 dB. The expected beneficial effect of the drug that caused this patient's symptoms is most likely due to which of the following actions? \textbf{A.} Inhibition of proteasome, \textbf{B.} Hyperstabilization of microtubules, \textbf{C.} Generation of free radicals, \textbf{D.} Cross-linking of DNA \\
        \midrule
        \rowcolor[RGB]{234, 238, 234}
        \bf \makecell[c]{CoT} \\
        \midrule
\textbf{Analysis}: The patient is experiencing a ringing sensation in his ear, which is a symptom of tinnitus.
[...]
\sethlcolor{lightred}\hl{Among the given options, the most likely mechanism of action that could cause sensorineural hearing loss is the generation of free radicals.} Free radicals can cause oxidative damage to the hair cells and auditory nerve, leading to hearing loss. Therefore, the answer choice is C. Generation of free radicals. \textbf{Answer}: C  \textcolor{red}{\xmark} \\
        \midrule
        \rowcolor[RGB]{234, 238, 234}
        \bf \makecell[c]{MedRAG} \\
        \midrule
\textbf{Retrieved documents}: [...]
\textbf{Analysis}:
[...]
To determine the mechanism of action of the drug causing the symptoms, we need to identify the drug used for neoadjuvant chemotherapy. However, \sethlcolor{lightred}\hl{the drug is not mentioned in the question stem or the provided documents. Without this information, it is not possible to definitively determine the mechanism of action.} Therefore, the answer cannot be determined based on the given information. \textbf{Answer}: Cannot be determined.  \textcolor{red}{\xmark} \\
        \midrule
        \rowcolor[RGB]{234, 238, 234}
        \bf \makecell[c]{\textit{i}-MedRAG} \\
        \midrule
\textit{\textbf{Query (Iteration 1)}}: What is the mechanism of action of the neoadjuvant chemotherapy drug used in transitional cell carcinoma of the bladder? \textbf{Retrieved documents}: [...] \\
\textit{\textbf{Answer}}: Based on the provided documents, the mechanism of action of the neoadjuvant chemotherapy drug used in transitional cell carcinoma of the bladder is not explicitly mentioned [...] \\
\textit{\textbf{Query (Iteration 2)}}: What are the potential side effects of neoadjuvant chemotherapy drugs used in transitional cell carcinoma of the bladder? \textbf{Retrieved documents}: [...] \\
\textit{\textbf{Answer}}: The potential side effects [...] include myelosuppression, dysuria, urinary frequency, contact dermatitis, and systemic illness associated with granulomatous infections in multiple sites.\\
\textit{\textbf{Query (Iteration 3)}}: What is the specific neoadjuvant chemotherapy drug
[...]
that can cause sensorineural hearing loss? \textbf{Retrieved documents}: [...] \\
\textit{\textbf{Answer}}: \hl{The specific neoadjuvant chemotherapy drug used in transitional cell carcinoma of the bladder that can cause sensorineural hearing loss is \textbf{cisplatin}}.\\
\textit{\textbf{Query (Iteration 4)}}: What is the \hl{mechanism of action of \textbf{cisplatin}?} \textbf{Retrieved documents}: [...] \\
\textit{\textbf{Answer}}: The mechanism of action of cisplatin is achieved through the covalent binding of platinum to the purine bases guanine and adenine in DNA. This covalent binding leads to \hl{intra-strand and inter-strand crosslinks} causing subsequent strand breaks [...] \\
\textbf{Final analysis}: [...] The drug that is most likely responsible for the patient's symptoms is cisplatin, which is known to cause sensorineural hearing loss. The mechanism of action of cisplatin involves cross-linking of DNA, leading to subsequent strand breaks and cell death. Therefore, the correct answer is D. \textbf{Answer}: D \greencheck\\
        \bottomrule
    \end{tabular}}
    \label{tab:case_study}
\end{table}

\begin{table}[h!] \small
    \tbl{An example of how \textit{i}-MedRAG helps Llama-3.1-8B to find the correct answer. The \sethlcolor{lightred}\hl{incorrect rationale} and \sethlcolor{yellow}\hl{useful evidence} are highlighted.}{
    \centering
    \begin{tabular}{p{16.5cm}}
        \toprule
        \bf \makecell[c]{MMLU Question} \\
        \midrule
\textbf{professional\_medicine-020}: A 52-year-old woman comes to the emergency department because of a 1-week history of low-grade fever and increasing abdominal cramps that are exacerbated by bowel movements [...] Her temperature is 38.1$^\circ$C (100.6$^\circ$F), pulse is 75/min, and blood pressure is 150/80 mm Hg [...] there is tenderness to palpation of the left lower quadrant with guarding but no rebound [...] leukocyte count is 15,000/mm3 [...] Which of the following is the most appropriate next step in diagnosis? \textbf{A.} Test of the stool for Clostridium difficile toxin \textbf{B.} Endoscopic retrograde cholangiopancreatography \textbf{C.} Pelvic ultrasonography \textbf{D.} CT scan of the abdomen with contrast\\
        \midrule
        \rowcolor[RGB]{234, 238, 234}
        \bf \makecell[c]{CoT} \\
        \midrule
\textbf{Analysis}: [...]
the differential diagnosis includes diverticulitis, appendicitis, and bowel obstruction. However, \sethlcolor{lightred}\hl{the absence of bloody stools and the patient's response to antibiotics makes diverticulitis less likely}. Therefore, the most appropriate next step in diagnosis is to test for Clostridium difficile toxin in the stool. \textbf{Answer}: A \textcolor{red}{\xmark} \\
        \midrule
        \rowcolor[RGB]{234, 238, 234}
        \bf \makecell[c]{MedRAG} \\
        \midrule
\textbf{Retrieved documents}: [...] \textbf{Analysis}: [...] Given the patient's symptoms and history, \sethlcolor{lightred}\hl{the most likely diagnosis is a pelvic abscess or a complication related to the previous surgery}. The most appropriate next step in diagnosis would be to investigate the pelvic area for any potential complications or abscesses. \textbf{Answer}: C \textcolor{red}{\xmark} \\
        \midrule
        \rowcolor[RGB]{234, 238, 234}
        \bf \makecell[c]{\textit{i}-MedRAG} \\
        \midrule
\textbf{\textit{Query 1 (Iteration 1)}}: What are the common symptoms and diagnostic criteria for Clostridium difficile infection (CDI)? \textbf{Retrieved documents}: [...]\\
\textbf{Answer}: 
[...] \\
\textbf{\textit{Query 2 (Iteration 1)}}: What are the typical findings on physical examination and laboratory tests for diverticulitis? \textbf{Retrieved documents}: [...]\\
\textbf{Answer}: Based on the provided documents, the typical findings on physical examination and laboratory tests for diverticulitis are: \textit{Physical Examination}: \hl{Tenderness to palpation over the area of inflammation (almost always present)} [...] \hl{Fever is almost always present}, but hypotension and shock are uncommon. \textit{Laboratory Tests}: \hl{Leukocytosis} (elevated white blood cell count) [...] \\
\textbf{\textit{Query 1 (Iteration 2)}}: What are the typical imaging findings for diverticulitis on a CT scan of the abdomen with contrast? \textbf{Retrieved documents}: [...]\\
\textbf{Answer}: [...] These findings are consistent with the diagnosis of diverticulitis, and \hl{are often used to differentiate it from other conditions that may present with similar symptoms.}\\
\textbf{\textit{Query 2 (Iteration 2)}}: What are the indications for surgical intervention in a patient with diverticulitis? \textbf{Retrieved documents}: [...]\\
\textbf{Answer}: 
[...] \\
\textbf{Final Analysis}:
[...]
The patient's laboratory results show a leukocyte count of 15,000/mm$^3$, indicating leukocytosis
[...]
The patient's symptoms of abdominal cramps, fever, and leukocytosis, combined with tenderness to palpation of the left lower quadrant, \hl{are consistent with diverticulitis} [...] the most appropriate next step in diagnosis is to perform a \hl{CT scan} of the abdomen with contrast to evaluate for diverticulitis and rule out other causes of her symptoms. \textbf{Answer}: D \greencheck\\
        \bottomrule
    \end{tabular}}
    \label{tab:case_study_mmlu}
\end{table}
Table \ref{tab:case_study} shows the predictions of GPT-3.5 on a MedQA question with different prompt engineering approaches.
The question asks about the mechanism of the drug for transitional cell carcinoma of the bladder which causes hearing loss. To solve the problem, it is important to find the exact drug and then figure out how it causes the mentioned symptoms.
However, the CoT result shows that GPT-3.5 does not inherently contain sufficient medical knowledge to solve this problem. Instead of inferring the described drug, GPT-3.5 with CoT directly hallucinates a wrong option as the answer. While free radicals are relevant to hearing loss, their connection to the disease of the patient is unclear and not discussed. Compared to CoT which solely relies on the internal knowledge of LLMs, MedRAG provides an opportunity for LLMs to augment their answer generation with external medical knowledge. Nevertheless, the model output shows that the MedRAG system fails to retrieve useful information about the drug from medical corpora. Given the complex problem description, it is difficult for text retrievers to find the asked mechanism without knowing the drug.

With iteratively generated follow-up queries, our \textit{i}-MedRAG manages to identify the described drug and find information about its mechanism. From Table \ref{tab:case_study}, it can be observed that GPT-3.5 starts with a general query about the asked mechanism. However, similar to the case in MedRAG, the RAG system fails to provide useful information about the query. With the information-seeking history, GPT-3.5 updates its actions with follow-up queries with respect to side effects especially hearing loss. With the updated queries, it manages to identify ``cisplatin'' as the drug which is not explicitly mentioned in the question. A query about the mechanism of action of cisplatin is further proposed to search for information about the answer to the original question. With several iterations of adaptive question answering, GPT-3.5 successfully finds the correct answer for the given clinical medical question.

Table \ref{tab:case_study_mmlu} shows another case from the MMLU-Med dataset by Llama-3.1-8B. The case is selected from the ``professional medicine'' subtask of MMLU-Med, which contains complex clinical cases similar to those in MedQA. In the question, LLMs are asked to choose the next step in diagnosis given the described symptoms, which requires the initial analysis of potential diseases. Similar to GPT-3.5, the open-source Llama-3.1-8B may hallucinate wrong answers with just CoT or MedRAG, as shown by the highlighted incorrect rationales in Table \ref{tab:case_study_mmlu}. With iterative follow-up queries, \textit{i}-MedRAG enables the LLM to find out the correct answer by making specific queries given existing information. For example, Llama-3.1-8B asks about the symptoms of \textit{Clostridium difficile} infection (CDI) and diverticulitis in its first iteration which helps it to identify diverticulitis as the potential disease for the diagnosis. In the second iteration, the model queries about the CT scan for diverticulitis, where the answer provides key information that helps it identify the correct next step.

\section{Discussion}
Overall, our proposed \textit{i}-MedRAG effectively improves the performance of LLMs on complex medical questions by prompting them to iteratively ask follow-up queries. The experimental results show that our approach is better than previously proposed prompt engineering and fine-tuning methods, and is generalizable to various LLMs and medical question-answering datasets. Nevertheless, our approach has certain limitations which need to be discussed. It is also worthwhile to discuss the future work of this study to analyze how it can be further improved to facilitate real-world medical assistance.

\subsection{Limitations}
The first limitation of \textit{i}-MedRAG is its high cost. While generating more follow-up queries tends to provide LLMs with more comprehensive and focused information about the given medical question, the cost also grows linearly with the number of queries generated. The time cost can be further increased if more documents are used to help answer the generated queries with RAG. While the cost is comparable to approaches using multiple LLM agents\cite{tang2023medagents} or self consistency\cite{wangself} which also prompt LLMs multiple times for each question, it is much more costly than baseline prompting methods such as CoT\cite{wei2022chain}.

Another limitation is the selection of hyperparameter values for optimal performance. As shown in Figure \ref{fig:scaling}, different LLMs can have different hyperparameter settings for their optimal performance. Even for the same LLM, its optimal hyperparameters can vary based on the medical questions being processed. Thus, it is non-trivial to find the optimal hyperparameters of \textit{i}-MedRAG for a new medical task, which may be inefficient for real-world deployments.

\subsection{Future work}

Given the limitations of \textit{i}-MedRAG, we consider several potential future directions that could further improve the performance of retrieval-augmented generation for medicine. The first direction is the automation of hyperparameter selection in \textit{i}-MedRAG. To reduce the laborious process of hyperparameter selection, one may use an LLM agent to dynamically determine how many follow-up queries should be asked each iteration. This can improve the efficiency and flexibility of the hyperparameter selection process. Another future direction is to improve the performance of \textit{i}-MedRAG with few-shot demonstrations. While few-shot CoT prompting is demonstrated to perform better than the zero-shot counterpart\cite{nori2023capabilities}, it is not easy to adapt such strategies to \textit{i}-MedRAG as the reasoning process can be dynamically affected by the use of external corpora and retrievers. Investigating how \textit{i}-MedRAG can benefit from one or few-shot samples could be a potential direction to further enhance its performance on medical question answering. More quantitative analysis can also be performed to examine the error types of \textit{i}-MedRAG compared to existing methods.

\section{Acknowledgments}
Guangzhi Xiong and Aidong Zhang are supported
by NIH grant 1R01LM014012 and NSF grant
2333740. Qiao Jin and Zhiyong Lu are supported by the Intramural Research Program of the National Library of Medicine, National Institutes of Health.

\bibliographystyle{ws-procs11x85}
\bibliography{ws-pro-sample}

\begin{thebibliography}{10}

\bibitem{thirunavukarasu2023large}
A.~J. Thirunavukarasu, D.~S.~J. Ting, K.~Elangovan, L.~Gutierrez, T.~F. Tan and D.~S.~W. Ting, Large language models in medicine, {\em Nature medicine} {\bf 29}, 1930  (2023).

\bibitem{zhou2023survey}
H.~Zhou, B.~Gu, X.~Zou, Y.~Li, S.~S. Chen, P.~Zhou, J.~Liu, Y.~Hua, C.~Mao, X.~Wu {\em et~al.}, A survey of large language models in medicine: Progress, application, and challenge, {\em arXiv preprint arXiv:2311.05112}   (2023).

\bibitem{omiye2024large}
J.~A. Omiye, H.~Gui, S.~J. Rezaei, J.~Zou and R.~Daneshjou, Large language models in medicine: the potentials and pitfalls: a narrative review, {\em Annals of Internal Medicine} {\bf 177}, 210  (2024).

\bibitem{tian2024opportunities}
S.~Tian, Q.~Jin, L.~Yeganova, P.-T. Lai, Q.~Zhu, X.~Chen, Y.~Yang, Q.~Chen, W.~Kim, D.~C. Comeau {\em et~al.}, Opportunities and challenges for chatgpt and large language models in biomedicine and health, {\em Briefings in Bioinformatics} {\bf 25}, p. bbad493  (2024).

\bibitem{luo2022biogpt}
R.~Luo, L.~Sun, Y.~Xia, T.~Qin, S.~Zhang, H.~Poon and T.-Y. Liu, Biogpt: generative pre-trained transformer for biomedical text generation and mining, {\em Briefings in bioinformatics} {\bf 23}, p. bbac409  (2022).

\bibitem{lievin2024can}
V.~Li{\'e}vin, C.~E. Hother, A.~G. Motzfeldt and O.~Winther, Can large language models reason about medical questions?, {\em Patterns} {\bf 5}  (2024).

\bibitem{nori2023can}
H.~Nori, Y.~T. Lee, S.~Zhang, D.~Carignan, R.~Edgar, N.~Fusi, N.~King, J.~Larson, Y.~Li, W.~Liu {\em et~al.}, Can generalist foundation models outcompete special-purpose tuning? case study in medicine, {\em arXiv preprint arXiv:2311.16452}   (2023).

\bibitem{singhal2023large}
K.~Singhal, S.~Azizi, T.~Tu, S.~S. Mahdavi, J.~Wei, H.~W. Chung, N.~Scales, A.~Tanwani, H.~Cole-Lewis, S.~Pfohl {\em et~al.}, Large language models encode clinical knowledge, {\em Nature} {\bf 620}, 172  (2023).

\bibitem{bolton2024biomedlm}
E.~Bolton, A.~Venigalla, M.~Yasunaga, D.~Hall, B.~Xiong, T.~Lee, R.~Daneshjou, J.~Frankle, P.~Liang, M.~Carbin {\em et~al.}, Biomedlm: A 2.7 b parameter language model trained on biomedical text, {\em arXiv preprint arXiv:2403.18421}   (2024).

\bibitem{shaib2023summarizing}
C.~Shaib, M.~Li, S.~Joseph, I.~Marshall, J.~J. Li and B.~C. Wallace, Summarizing, simplifying, and synthesizing medical evidence using gpt-3 (with varying success), 1387  (2023).

\bibitem{tang2023evaluating}
L.~Tang, Z.~Sun, B.~Idnay, J.~G. Nestor, A.~Soroush, P.~A. Elias, Z.~Xu, Y.~Ding, G.~Durrett, J.~F. Rousseau {\em et~al.}, Evaluating large language models on medical evidence summarization, {\em npj Digital Medicine} {\bf 6}, p. 158  (2023).

\bibitem{van2024adapted}
D.~Van~Veen, C.~Van~Uden, L.~Blankemeier, J.-B. Delbrouck, A.~Aali, C.~Bluethgen, A.~Pareek, M.~Polacin, E.~P. Reis, A.~Seehofnerov{\'a} {\em et~al.}, Adapted large language models can outperform medical experts in clinical text summarization, {\em Nature Medicine} , 1  (2024).

\bibitem{jin2023matching}
Q.~Jin, Z.~Wang, C.~S. Floudas, F.~Chen, C.~Gong, D.~Bracken-Clarke, E.~Xue, Y.~Yang, J.~Sun and Z.~Lu, Matching patients to clinical trials with large language models, {\em ArXiv}   (2023).

\bibitem{wong2023scaling}
C.~Wong, S.~Zhang, Y.~Gu, C.~Moung, J.~Abel, N.~Usuyama, R.~Weerasinghe, B.~Piening, T.~Naumann, C.~Bifulco {\em et~al.}, Scaling clinical trial matching using large language models: A case study in oncology, 846  (2023).

\bibitem{wornow2024zero}
M.~Wornow, A.~Lozano, D.~Dash, J.~Jindal, K.~W. Mahaffey and N.~H. Shah, Zero-shot clinical trial patient matching with llms, {\em arXiv preprint arXiv:2402.05125}   (2024).

\bibitem{zhuang2024team}
S.~Zhuang, B.~Koopman and G.~Zuccon, Team ielab at trec clinical trial track 2023: Enhancing clinical trial retrieval with neural rankers and large language models, {\em arXiv preprint arXiv:2401.01566}   (2024).

\bibitem{ji2023survey}
Z.~Ji, N.~Lee, R.~Frieske, T.~Yu, D.~Su, Y.~Xu, E.~Ishii, Y.~J. Bang, A.~Madotto and P.~Fung, Survey of hallucination in natural language generation, {\em ACM Computing Surveys} {\bf 55}, 1  (2023).

\bibitem{wu2024continual}
T.~Wu, L.~Luo, Y.-F. Li, S.~Pan, T.-T. Vu and G.~Haffari, Continual learning for large language models: A survey, {\em arXiv preprint arXiv:2402.01364}   (2024).

\bibitem{lewis2020retrieval}
P.~Lewis, E.~Perez, A.~Piktus, F.~Petroni, V.~Karpukhin, N.~Goyal, H.~K{\"u}ttler, M.~Lewis, W.-t. Yih, T.~Rockt{\"a}schel {\em et~al.}, Retrieval-augmented generation for knowledge-intensive nlp tasks, {\em Advances in Neural Information Processing Systems} {\bf 33}, 9459  (2020).

\bibitem{gao2023retrieval}
Y.~Gao, Y.~Xiong, X.~Gao, K.~Jia, J.~Pan, Y.~Bi, Y.~Dai, J.~Sun and H.~Wang, Retrieval-augmented generation for large language models: A survey, {\em arXiv preprint arXiv:2312.10997}   (2023).

\bibitem{zakka2024almanac}
C.~Zakka, R.~Shad, A.~Chaurasia, A.~R. Dalal, J.~L. Kim, M.~Moor, R.~Fong, C.~Phillips, K.~Alexander, E.~Ashley {\em et~al.}, Almanac—retrieval-augmented language models for clinical medicine, {\em NEJM AI} {\bf 1}, p. AIoa2300068  (2024).

\bibitem{lozano2023clinfo}
A.~Lozano, S.~L. Fleming, C.-C. Chiang and N.~Shah, Clinfo. ai: An open-source retrieval-augmented large language model system for answering medical questions using scientific literature, 8  (2023).

\bibitem{xiong2024benchmarking}
G.~Xiong, Q.~Jin, Z.~Lu and A.~Zhang, Benchmarking retrieval-augmented generation for medicine, {\em arXiv preprint arXiv:2402.13178}   (2024).

\bibitem{jin2019pubmedqa}
Q.~Jin, B.~Dhingra, Z.~Liu, W.~Cohen and X.~Lu, Pubmedqa: A dataset for biomedical research question answering, 2567  (2019).

\bibitem{tsatsaronis2015overview}
G.~Tsatsaronis, G.~Balikas, P.~Malakasiotis, I.~Partalas, M.~Zschunke, M.~R. Alvers, D.~Weissenborn, A.~Krithara, S.~Petridis, D.~Polychronopoulos {\em et~al.}, An overview of the bioasq large-scale biomedical semantic indexing and question answering competition, {\em BMC bioinformatics} {\bf 16}, 1  (2015).

\bibitem{jin2021disease}
D.~Jin, E.~Pan, N.~Oufattole, W.-H. Weng, H.~Fang and P.~Szolovits, What disease does this patient have? a large-scale open domain question answering dataset from medical exams, {\em Applied Sciences} {\bf 11}, p. 6421  (2021).

\bibitem{low2024answering}
Y.~S. Low, M.~L. Jackson, R.~J. Hyde, R.~E. Brown, N.~M. Sanghavi, J.~D. Baldwin, C.~W. Pike, J.~Muralidharan, G.~Hui, N.~Alexander {\em et~al.}, Answering real-world clinical questions using large language model based systems, {\em arXiv preprint arXiv:2407.00541}   (2024).

\bibitem{jin2024pubmed}
Q.~Jin, R.~Leaman and Z.~Lu, Pubmed and beyond: biomedical literature search in the age of artificial intelligence, {\em EBioMedicine} {\bf 100}  (2024).

\bibitem{jin2023retrieve}
Q.~Jin, R.~Leaman and Z.~Lu, Retrieve, summarize, and verify: how will chatgpt affect information seeking from the medical literature?, {\em Journal of the American Society of Nephrology} {\bf 34}, 1302  (2023).

\bibitem{li2024self}
X.~Li, P.~Yu, C.~Zhou, T.~Schick, O.~Levy, L.~Zettlemoyer, J.~E. Weston and M.~Lewis, Self-alignment with instruction backtranslation  (2024).

\bibitem{shao2023enhancing}
Z.~Shao, Y.~Gong, Y.~Shen, M.~Huang, N.~Duan and W.~Chen, Enhancing retrieval-augmented large language models with iterative retrieval-generation synergy, 9248  (2023).

\bibitem{trivedi2023interleaving}
H.~Trivedi, N.~Balasubramanian, T.~Khot and A.~Sabharwal, Interleaving retrieval with chain-of-thought reasoning for knowledge-intensive multi-step questions, 10014  (2023).

\bibitem{jiang2024retrieve}
Z.~Jiang, M.~Sun, L.~Liang and Z.~Zhang, Retrieve, summarize, plan: Advancing multi-hop question answering with an iterative approach, {\em arXiv preprint arXiv:2407.13101}   (2024).

\bibitem{pal2022medmcqa}
A.~Pal, L.~K. Umapathi and M.~Sankarasubbu, Medmcqa: A large-scale multi-subject multi-choice dataset for medical domain question answering, 248  (2022).

\bibitem{hendrycks2020measuring}
D.~Hendrycks, C.~Burns, S.~Basart, A.~Zou, M.~Mazeika, D.~Song and J.~Steinhardt, Measuring massive multitask language understanding, {\em arXiv preprint arXiv:2009.03300}   (2020).

\bibitem{jin2022biomedical}
Q.~Jin, Z.~Yuan, G.~Xiong, Q.~Yu, H.~Ying, C.~Tan, M.~Chen, S.~Huang, X.~Liu and S.~Yu, Biomedical question answering: a survey of approaches and challenges, {\em ACM Computing Surveys (CSUR)} {\bf 55}, 1  (2022).

\bibitem{wei2022chain}
J.~Wei, X.~Wang, D.~Schuurmans, M.~Bosma, F.~Xia, E.~Chi, Q.~V. Le, D.~Zhou {\em et~al.}, Chain-of-thought prompting elicits reasoning in large language models, {\em Advances in neural information processing systems} {\bf 35}, 24824  (2022).

\bibitem{wang2022self}
X.~Wang, J.~Wei, D.~Schuurmans, Q.~Le, E.~Chi, S.~Narang, A.~Chowdhery and D.~Zhou, Self-consistency improves chain of thought reasoning in language models, {\em arXiv preprint arXiv:2203.11171}   (2022).

\bibitem{tang2023medagents}
X.~Tang, A.~Zou, Z.~Zhang, Y.~Zhao, X.~Zhang, A.~Cohan and M.~Gerstein, Medagents: Large language models as collaborators for zero-shot medical reasoning, {\em arXiv preprint arXiv:2311.10537}   (2023).

\bibitem{feng2023knowledge}
C.~Feng, X.~Zhang and Z.~Fei, Knowledge solver: Teaching llms to search for domain knowledge from knowledge graphs, {\em arXiv preprint arXiv:2309.03118}   (2023).

\bibitem{wang2023augmenting}
Y.~Wang, X.~Ma and W.~Chen, Augmenting black-box llms with medical textbooks for clinical question answering, {\em arXiv preprint arXiv:2309.02233}   (2023).

\bibitem{savage2024diagnostic}
T.~Savage, A.~Nayak, R.~Gallo, E.~Rangan and J.~H. Chen, Diagnostic reasoning prompts reveal the potential for large language model interpretability in medicine, {\em NPJ Digital Medicine} {\bf 7}, p.~20  (2024).

\bibitem{nori2023capabilities}
H.~Nori, N.~King, S.~M. McKinney, D.~Carignan and E.~Horvitz, Capabilities of gpt-4 on medical challenge problems, {\em arXiv preprint arXiv:2303.13375}   (2023).

\bibitem{saab2024capabilities}
K.~Saab, T.~Tu, W.-H. Weng, R.~Tanno, D.~Stutz, E.~Wulczyn, F.~Zhang, T.~Strother, C.~Park, E.~Vedadi {\em et~al.}, Capabilities of gemini models in medicine, {\em arXiv preprint arXiv:2404.18416}   (2024).

\bibitem{shi2024medadapter}
W.~Shi, R.~Xu, Y.~Zhuang, Y.~Yu, H.~Wu, C.~Yang and M.~D. Wang, Medadapter: Efficient test-time adaptation of large language models towards medical reasoning, {\em arXiv preprint arXiv:2405.03000}   (2024).

\bibitem{jin2023medcpt}
Q.~Jin, W.~Kim, Q.~Chen, D.~C. Comeau, L.~Yeganova, W.~J. Wilbur and Z.~Lu, Medcpt: Contrastive pre-trained transformers with large-scale pubmed search logs for zero-shot biomedical information retrieval, {\em Bioinformatics} {\bf 39}, p. btad651  (2023).

\bibitem{wangself}
X.~Wang, J.~Wei, D.~Schuurmans, Q.~V. Le, E.~H. Chi, S.~Narang, A.~Chowdhery and D.~Zhou, Self-consistency improves chain of thought reasoning in language models.

\end{thebibliography}

\end{document}